\documentclass[11pt,a4paper]{article}
\usepackage[dvipsnames]{xcolor}
\usepackage[hyperref]{naaclhlt2018}
\usepackage{mathptmx} 
\usepackage{latexsym}
\usepackage{amsfonts}
\usepackage{amsmath}

\usepackage{url}
\usepackage{listings}
\usepackage{microtype}
\usepackage{array}
\usepackage{booktabs}
\usepackage{todonotes}
\usepackage{pgfplots}
\usepackage{linguex}
\usepackage[inline]{enumitem} 
\usepackage[group-separator={,}]{siunitx}
\usepackage{tikz4nlp}
\aclfinalcopy 

\newcommand{\tagroundedcorners}{0.75ex}
\newcommand{\mentiontag}[2]{\tikz[remember picture,baseline=-0.5ex]{\node [draw=none,inner sep=0.75ex,rounded corners=\tagroundedcorners,fill=cb_lightgray,minimum size=4ex,text width={}] (#2) {#1};}}

\title{Mixing Context Granularities for Improved Entity Linking \\ on Question Answering Data across Entity Categories}

\author{Daniil Sorokin \and Iryna Gurevych\\
Ubiquitous Knowledge Processing Lab (UKP) \and Research Training Group AIPHES \\
Department of Computer Science, Technische Universit\"at Darmstadt \\
  {\url{www.ukp.tu-darmstadt.de}} \\ 
}

\date{}

\begin{document}
\maketitle

\begin{abstract}
The first stage of every knowledge base question answering approach is to link entities in the input question. 
We investigate entity linking in the context of a question answering task and present a jointly optimized neural architecture for entity mention detection and entity disambiguation that models the surrounding context on different levels of granularity. 
  
We use the Wikidata knowledge base and available question answering datasets to create benchmarks for entity linking on question answering data. 
Our approach outperforms the previous state-of-the-art system on this data, resulting in an average 8\% improvement of the final score. We further demonstrate that our model delivers a strong performance across different entity categories.

\end{abstract}

\section{Introduction}

\begin{figure}[t]
  \begin{tikzpicture}[
    remember picture,
    line width=0.25mm,
  ]
  \node (wikipediasnippet) [draw=gray,dotted,inner sep=1.5ex,fill=lightgray!10,text width=0.9\linewidth] {
  \texttt{
    what are \mentiontag{taylor swift's}{entity1}  \mentiontag{albums}{entity2}?
  }
  };
  \matrix[matrix of nodes,
    line width=0.25mm,
    ampersand replacement=\&,	
    nodes={inner sep=1ex,draw=cb_darkgreen,fill=none,rounded corners=1ex,execute at begin node=\sffamily},
    column sep=0.5em,
    below=0.5ex of wikipediasnippet
    ] (wikidatids) {
  Taylor Swift Q462 \& album Q24951125\\
  };
  \begin{scope}[draw=cb_darkgreen!75,dash pattern=on 3pt off 3pt]
  \draw[<-,rel_arrow,line width=0.25mm] (wikidatids-1-1) -- (entity1);
  \draw[<-,rel_arrow,line width=0.25mm] (wikidatids-1-2) -- (entity2);
  \end{scope}

  \node (answer) [inner sep=1ex,draw=cb_darkyellow,fill=none,rounded corners=1ex,execute at begin node=\sffamily,xshift=6ex, below=11ex of wikipediasnippet] {Red, 1989, etc.};
  \path
  (wikidatids-1-1) edge [->,rel_arrow,bend right=30] node[text_above_arrow,midway,xshift=-2em,yshift=
  0.5ex] {performer} (answer.west); 
  \path
  (wikidatids-1-2) edge [->,rel_arrow,bend left=10] node[text_above_arrow,midway,xshift=2em,yshift=
0.5ex] {instance of} (answer.north);

  \end{tikzpicture}
  \caption{An example question from a QA dataset that shows the correct entity mentions and their relationship with the correct answer to the question, {\sffamily Qxxx} stands for a knowledge base identifier 
  \label{fig:example-question}}
\end{figure}
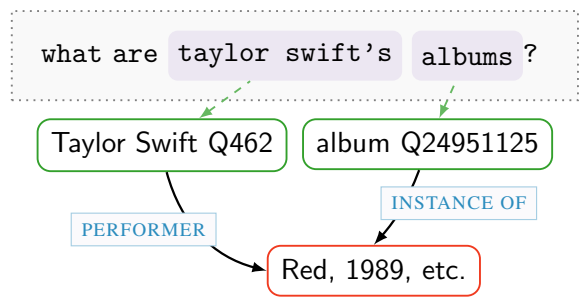

Knowledge base question answering (QA) requires a precise modeling of the question semantics through the entities and relations available in the knowledge base (KB) in order to retrieve the correct answer.
The first stage for every QA approach is entity linking (EL), that is the identification of entity mentions in the question and linking them to entities in KB. In Figure~\ref{fig:example-question}, two entity mentions are detected and linked to the knowledge base referents. This step is crucial for QA since the correct answer must be connected via some path over KB to the entities mentioned in the question. 

The state-of-the-art QA systems usually rely on off-the-shelf EL systems to extract entities from the question~\cite{Yih2015}.
Multiple EL systems are freely available and can be readily applied for question answering (e.g. DBPedia Spotlight\footnote{\url{http://www.dbpedia-spotlight.org}}, AIDA\footnote{\url{https://www.mpi-inf.mpg.de/yago-naga/aida/}}). However, these systems have certain drawbacks in the QA setting: they are targeted at long well-formed documents, such as news texts, and are less suited for typically short and noisy question data. 
Other EL systems focus on noisy data (e.g. S-MART, \citealp{Yang2015a}), but are not openly available and hence limited in their usage and application.
Multiple error analyses of QA systems point to entity linking as a major external source of error~\cite{Berant2014,Reddy2014,Yih2015}.

The QA datasets are normally collected from the web and contain very noisy and diverse data~\cite{Berant2013}, which poses a number of challenges for EL. First, many common features used in EL systems, such as capitalization, are not meaningful on noisy data.
Moreover, a question is a short text snippet that does not contain broader context that is helpful for entity disambiguation. The QA data also features many entities of various categories and differs in this respect from the Twitter datasets that are often used to evaluate EL systems.

In this paper, we present an approach that tackles the challenges listed above: we perform entity mention detection and entity disambiguation jointly in a single neural model that makes the whole process end-to-end differentiable. This ensures that any token n-gram can be considered as a potential entity mention, which is important to be able to link entities of different categories, such as movie titles and organization names.

To overcome the noise in the data, we automatically learn features over a set of contexts of different granularity levels.
Each level of granularity is handled by a separate component of the model. A token-level component extracts higher-level features from the whole question context, whereas a character-level component builds lower-level features for the candidate n-gram. Simultaneously, we extract features from the knowledge base context of the candidate entity: character-level features are extracted for the entity label and higher-level features are produced based on the entities surrounding the candidate entity in the knowledge graph. This information is aggregated and used to predict whether the n-gram is an entity mention and to what entity it should be linked.

\textbf{Contributions} The two main contributions of our work are:
  \begin{enumerate}[label=(\roman*)]
    \item We construct two datasets to evaluate EL for QA and present a set of strong baselines: the existing EL systems that were used as a building block for QA before and a model that uses manual features from the previous work on noisy data.
    \item We design and implement an entity linking system that models contexts of variable granularity to detect and disambiguate entity mentions. To the best of our knowledge, we are the first to present a unified end-to-end neural model for entity linking for noisy data that operates on different context levels and does not rely on manual features.
    Our architecture addresses the challenges of entity linking on question answering data and outperforms state-of-the-art EL systems.
  \end{enumerate}

  \textbf{Code and datasets} Our system can be applied on any QA dataset. 
	The complete code as well as the scripts that produce the evaluation data can be found here: \url{https://github.com/UKPLab/starsem2018-entity-linking}.

\section{Motivation and Related Work}

Several benchmarks exist for EL on Wikipedia texts and news articles, such as ACE~\cite{Bentivogli2010} and CoNLL-YAGO~\cite{Hoffart2011a}. These datasets contain multi-sentence documents and largely cover three types of entities: Location, Person and Organization. These types are commonly recognized by named entity recognition systems, such as Stanford NER Tool~\cite{Manning2014}. Therefore in this scenario, an EL system can solely focus on entity disambiguation.

\begin{figure}[t]
  \begin{tikzpicture}
  \begin{axis}[
      ybar,
    height=7cm,
    width=\linewidth,    
      enlarge y limits=0.15,
      bar width=0.75ex,
      legend style={at={(0.5,-0.3)},
        anchor=north,legend columns=-1},
      ylabel={ratio of the entity type},
      symbolic x coords={event,location,organization,person,fic. character,product,thing,professions etc.},
      xtick=data,
      ymax=0.3,ymin=0.05,
    nodes near coords,
    nodes near coords align={vertical},
    every node near coord/.append style={font=\tiny, /pgf/number format/fixed,rotate=90,anchor=west},
      x tick label style={rotate=30,anchor=east,font=\small,text width=2cm, align=right},
      y tick label style={font=\small}, 
      ]
  \addplot[cb_darkblue,fill=cb_blue,pattern=north west lines,pattern color=cb_blue] coordinates {(fic. character, 0.009108653220559532)
   (event, 0.07937540663630449)
   (location, 0.2192582953806116)
   (organization, 0.24788549121665582)
   (professions etc., 0.1561483409238777)
   (person, 0.2225113858165257)
   (product, 0.05074821080026025)
   (thing, 0.014964216005204945)
  };
  \addplot[cb_darkgreen!80,fill=cb_lightgreen!30] coordinates {
(fic. character, 0.03721444362564481)
   (event, 0.06705969049373618)
   (location, 0.2969786293294031)
   (organization,  0.07663964627855564)
   (professions etc., 0.05821665438467207)
   (person, 0.39572586588061903)
   (product, 0.045689019896831246)
   (thing, 0.02247605011053795)
  };
  \addplot[cb_darkyellow,fill=cb_yellow,pattern=crosshatch,pattern color=cb_darkyellow] coordinates {
  (fic. character, 0.03352984524686809)
  (event, 0.14885777450257923)
  (location, 0.13964627855563744)
  (organization, 0.078481945467944)
  (professions etc., 0.23470891672807664)
  (person, 0.13264554163596168)
  (product, 0.05747973470891673)
  (thing, 0.017317612380250553)
  };
  \node [red, right,xshift=1ex, cb_darkgreen!80,font=\tiny,rotate=90] at (axis cs: person,0.31) {0.4};
  
  \legend{NEEL, WebQSP, GraphQuestions}
  \end{axis}
  \end{tikzpicture}
  \caption{Distribution of entity categories in the NEEL 2014, WebQSP  and GraphQuestions datasets \label{fig:types-dist}}
  \end{figure}
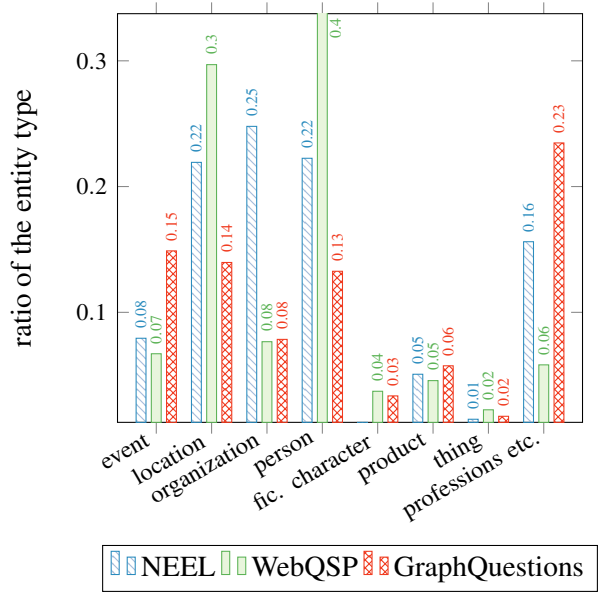

In the recent years, EL on Twitter data has emerged as a branch of entity linking research. In particular, EL on tweets was the central task of the NEEL shared task from 2014 to 2016~\cite{Hotho2016}. Tweets share some of the challenges with QA data: in both cases the input data is short and noisy. On the other hand, it significantly differs with respect to the entity types covered. The data for the NEEL shared task was annotated with 7 broad entity categories, that besides Location, Organization and Person include Fictional Characters, Events, Products (such as electronic devices or works of art) and Things (abstract objects). Figure~\ref{fig:types-dist} shows the distribution of entity categories in the training set from the NEEL 2014 competition. One can see on the diagram that the distribution is mainly skewed towards 3 categories: Location, Person and Organization. 

Figure~\ref{fig:types-dist} also shows the entity categories present in two QA datasets. The distribution over the categories is more diverse in this case. The WebQuestions dataset includes the Fictional Character and Thing categories which are almost absent from the NEEL dataset. A more even distribution can be observed in the GraphQuestion dataset that features many Events, Fictional Characters and Professions. This means that a successful system for EL on question data needs to be able to recognize and to link all categories of entities. Thus, we aim to show that comprehensive modeling of different context levels will result in a better generalization and performance across various entity categories.

\textbf{Existing Solutions} 
The early machine learning approaches to EL focused on long well-formed documents~\cite{Bunescu2006,Cucerzan2007,Han2012,Francis-Landau2016}. These systems usually rely on an off-the-shelf named entity recognizer to extract entity mentions in the input. 
As a consequence, such approaches can not handle entities of types other than those that are supplied by the named entity recognizer.
Named entity recognizers are normally trained to detect mentions of Locations, Organizations and Person names, whereas in the context of QA, the system also needs to cover movie titles, songs, common nouns such as `president' etc. 

To mitigate this, \citet{Cucerzan2012} has introduced the idea to perform mention detection and entity linking jointly using a linear combination of manually defined features. \citet{Luo2015a} have adopted the same idea and suggested a probabilistic graphical model for the joint prediction.
This is essential for linking entities in questions. For example in ``\textit{who does maggie grace play in taken?}'', it is hard to distinguish between the usage of the word `taken' and the title of a movie `Taken' without consulting a knowledge base.

\citet{Sun2015a} were among the first to use neural networks to embed the mention and the entity for a better prediction quality. Later,  \citet{Francis-Landau2016} have employed convolutional neural networks to extract features from the document context and mixed them with manually defined features, though they did not integrate it with mention detection.
\citet{Sil2017} continued the work in this direction recently and applied convolutional neural networks to cross-lingual EL.

The approaches that were developed for Twitter data present the most relevant work for EL on QA data. \citet{Guo2013a} have created a new dataset of around 1500 tweets and suggested a Structured SVM approach that handled mention detection and entity disambiguation together. \citet{Chang2014} describe the winning system of the NEEL 2014 competition on EL for short texts: The system adapts a joint approach similar to \citet{Guo2013a}, but uses the MART gradient boosting algorithm instead of the SVM and extends the feature set.
The current state-of-the-art system for EL on noisy data is S-MART~\citep{Yang2015a} which extends the approach from \citet{Chang2014} to make structured predictions. The same group has subsequently applied S-MART to extract entities for a QA system~\citep{Yih2015}.

Unfortunately, the described EL systems for short texts are not available as stand-alone tools. Consequently, the modern QA approaches mostly rely on off-the-shelf entity linkers that were designed for other domains.
\citet{Reddy2016} have employed the Freebase online API that was since deprecated. 
A number of question answering systems have relied on DBPedia Spotlight to extract entities~\cite{Lopez2016,Chen2016}. DBPedia Spotlight~\cite{Mendes2011} uses document similarity vectors, word embeddings and manually defined features such as entity frequency.
We are addressing this problem in our work by presenting an architecture specifically targeted at EL for QA data.

\textbf{The Knowledge Base} 
Throughout the experiments, we use the Wikidata\footnote{ At the moment, Wikidata contains more than 40 million entities and 350 million relation instances: \\ \raggedright \url{https://www.wikidata.org/wiki/Special:Statistics}} open-domain KB~\cite{Vrandecic2014}. 
Among the previous work, the common choices of a KB include Wikipedia, DBPedia and Freebase. The entities in Wikidata directly correspond to the Wikipedia articles, which enables us to work with data that was previously annotated with  DBPedia.
Freebase was discontinued and is no longer up-to-date. However, most entities in Wikidata have been annotated with identifiers from other knowledge sources and databases, including Freebase, which establishes a link between the two KBs.

\section{Entity Linking Architecture}
\label{sec:architecture}

\begin{figure}[ht]
  \centering{
  \includegraphics[width=0.95\linewidth]{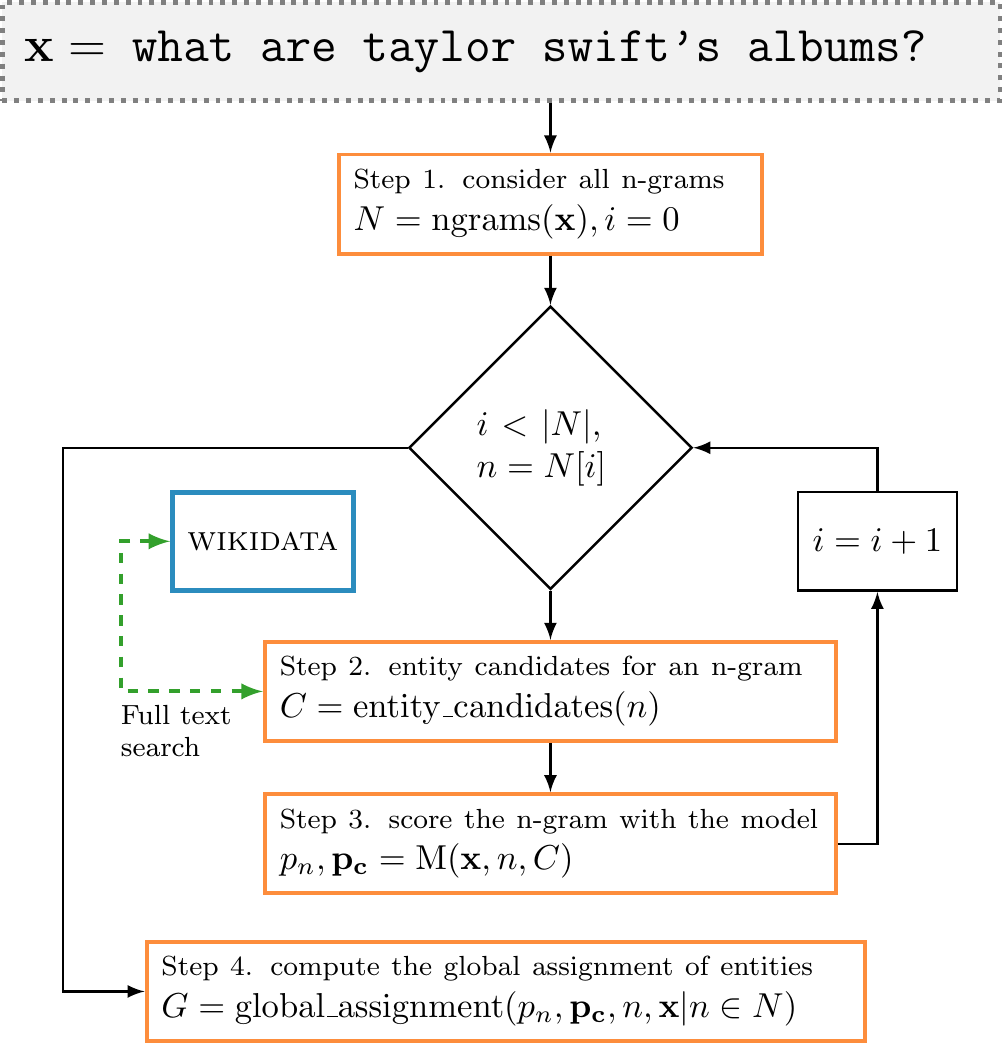}}
  \caption{Architecture of the entity linking system}
  \label{fig:system-diagram}
\end{figure}

The overall architecture of our entity linking system is depicted in Figure~\ref{fig:system-diagram}. From the input question $\mathbf{x}$ we extract all possible token n-grams $N$ up to a certain length as entity mention candidates~(Step 1). For each n-gram $n$, we look it up in the knowledge base using a full text search over entity labels (Step 2). That ensures that we find all entities that contain the given n-gram in the label. For example for a unigram `obama',  we retrieve `Barack Obama', `Michelle Obama' etc. 
This step produces a set of entity disambiguation candidates $C$ for the given n-gram $n$. We sort the retrieved candidates by length and cut off after the first $1000$. 
That ensures that the top candidates in the list would be those that exactly match the target n-gram $n$. 

In the next step, the list of n-grams $N$ and the corresponding list of entity disambiguation candidates are sent to the entity linking model (Step 3). 
The model jointly performs the detection of correct mentions and the disambiguation of entities. 

\subsection{Variable Context Granularity Network}

\begin{figure*}[t]
  {\centering
   \includegraphics[width=0.99\linewidth]{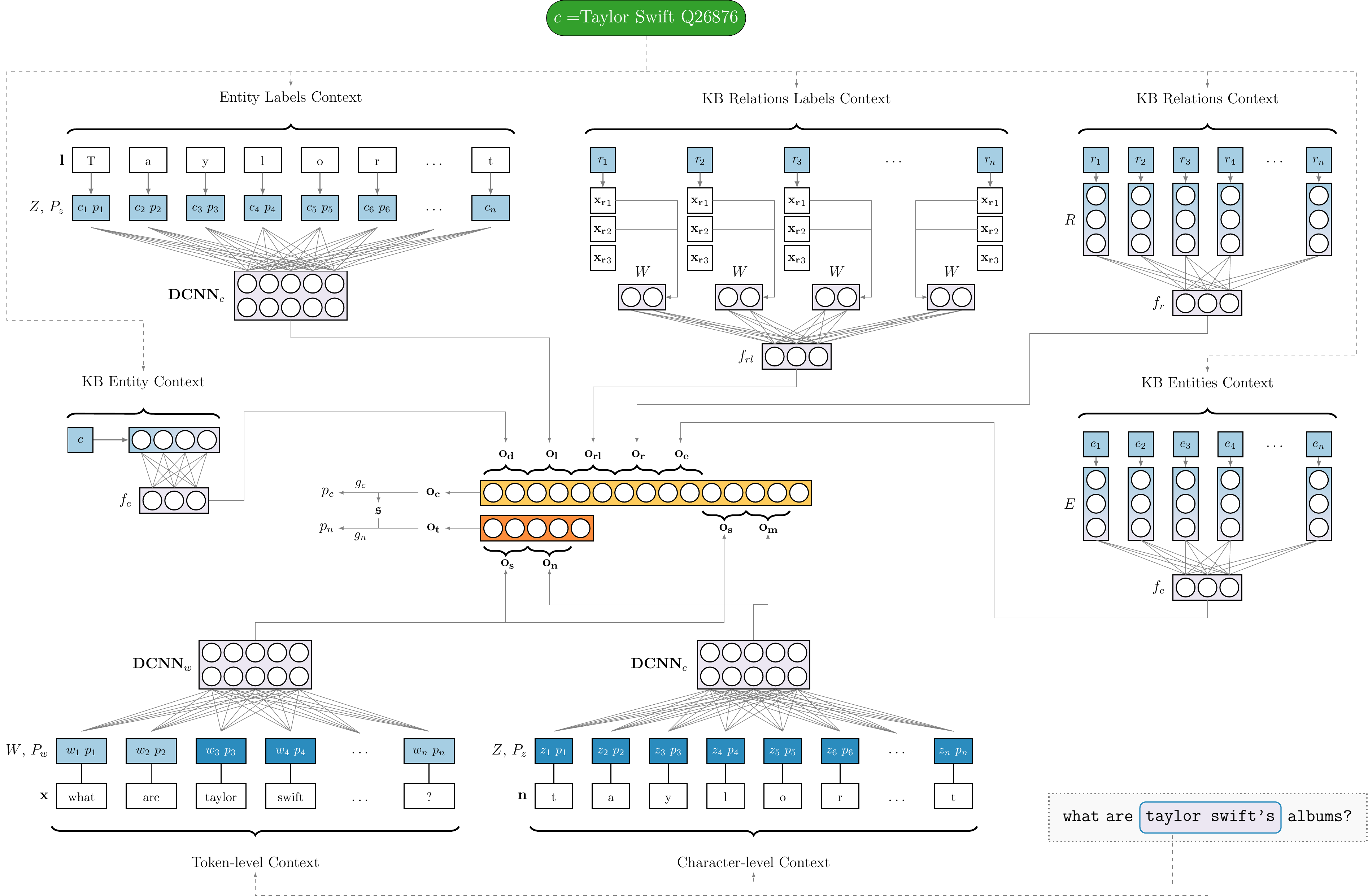}}
  \caption{The architecture of the Variable Context Granularity Network for a \textit{single} n-gram and an entity candidate. The output vectors $(\mathbf{o_c}, \mathbf{o_t})$ are aggregated over \textit{all} n-grams for the global assignment \label{fig:vcg-diagram}}
\end{figure*}

The neural architecture (Variable Context Granularity, VCG) aggregates and mixes contexts of different granularities to perform a joint mention detection and entity disambiguation. 
Figure~\ref{fig:vcg-diagram} shows the layout of the network and its main components.
The input to the model is a list of question tokens $\mathbf{x}$, a token n-gram $n$ and a list of candidate entities $C$. Then the model is a function $\mathrm{M}(\mathbf{x},n,C)$ that produces a mention detection score $p_n$ for each n-gram and a ranking score $p_c$ for each of the candidates $c \in C$: $p_n, \mathbf{p_c} = \mathrm{M}(\mathbf{x},n,C)$.

\textbf{Dilated Convolutions} To process sequential input, we use dilated convolutional networks (DCNN). \citet{Strubell2017} have recently shown that DCNNs are faster and as effective as recurrent models on the task of named entity recognition. We define two modules: $\mathbf{DCNN}_w$ and $\mathbf{DCNN}_c$ for processing token-level and character-level input respectively. Both modules consist of a series of convolutions applied with an increasing dilation, as described in \citet{Strubell2017}. The output of the convolutions is averaged and transformed by a fully-connected layer.

\textbf{Context components} The \textit{token component} corresponds to sentence-level features normally defined for EL and encodes the list of question tokens $\mathbf{x}$ into a fixed size vector. It maps the tokens in $\mathbf{x}$ to $d_w$-dimensional pre-trained word embeddings, using a matrix $\mathbf{W} \in \mathbb{R}^{|V_w| \times d_w}$, where $|V_w|$ is the size of the vocabulary. We use 50-dimensional GloVe embeddings pre-trained on a 6 billion tokens corpus~\cite{Pennington2014}. 
The word embeddings are concatenated with $d_p$-dimensional position embeddings $\mathbf{P_w} \in \mathbb{R}^{3 \times d_p}$ that are used to denote the tokens that are part of the target n-gram. The concatenated embeddings are processed by $\mathbf{DCNN}_w$ to get a vector $\mathbf{o_s}$.

\textit{The character component} processes the target token n-gram $n$ on the basis of individual characters. We add one token on the left and on the right to the target mention and map the string of characters to $d_z$-character embeddings, $\mathbf{Z} \in \mathbb{R}^{|V_z| \times d_z}$. We concatenate the character embeddings with $d_p$-dimensional position embeddings $\mathbf{P_z} \in \mathbb{R}^{|x| \times d_p}$ and process them with $\mathbf{DCNN}_c$ to get a feature vector $\mathbf{o_n}$.

We use \textit{the character component} with the same learned parameters to encode the label of a candidate entity from the KB as a vector $\mathbf{o_l}$. The parameter sharing between mention encoding and entity label encoding ensures that the representation of a mention is similar to the entity label.

The KB structure is the highest context level included in the model. \textit{The knowledge base structure component} models the entities and relations that are connected to the candidate entity $c$. First, we map a list of relations $\mathbf{r}$ of the candidate entity to $d_r$-dimensional pre-trained relations embeddings, using a matrix $\mathbf{R} \in \mathbb{R}^{|V_r| \times d_r}$, where $|V_r|$ is the number of relation types in the KB. We transform the relations embeddings with a single fully-connected layer $f_r$ and then apply a max pooling operation to get a single relation vector $\mathbf{o_r}$ per entity. Similarly, we map a list of entities that are immediately connected to the candidate entity $\mathbf{e}$ to $d_e$-dimensional pre-trained entity embeddings, using a matrix $\mathbf{E} \in \mathbb{R}^{|V_e| \times d_e}$, where $|V_e|$ is the number of entities in the KB. The entity embeddings are transformed by a fully-connected layer $f_e$ and then also pooled to produce the output $\mathbf{o_e}$. The embedding of the candidate entity itself is also transformed with $f_e$ and is stored as $\mathbf{o_d}$. To train the knowledge base embeddings, we use the TransE algorithm~\citep{Bordes2013}.

Finally, \textit{the knowledge base lexical component} takes the labels of the relations in $\mathbf{r}$ to compute lexical relation embeddings. For each $r \in \mathbf{r}$, we tokenize the label and map the tokens $\mathbf{x_r}$ to word embeddings, using the word embedding matrix $\mathbf{W}$. To get a single lexical embedding per relation, we apply max pooling and transform the output with a fully-connected layer $f_{rl}$. The lexical relation embeddings for the candidate entity are pooled into the vector $\mathbf{o_{rl}}$.

\textbf{Context Aggregation} The different levels of context are aggregated and are transformed by a sequence of fully-connected layers into a final vector $\mathbf{o_c}$ for the n-gram $n$ and the candidate entity $c$.
The vectors for each candidate are aggregated into a matrix $O = [\mathbf{o_c}| c \in C]$. We apply element-wise max pooling on $O$ to get a single summary vector $\mathfrak{s}$ for all entity candidates for $n$. 

To get the ranking score $p_c$ for each entity candidate $c$, we apply a single fully-connected layer $g_c$ on the concatenation of $\mathbf{o_c}$ and the summary vector $\mathfrak{s} $: $p_c = g_c(\mathbf{o_c} \| \mathfrak{s} )$.
For the mention detection score for the n-gram, we separately concatenate the vectors for the token context $\mathbf{o_s}$ and the character context $\mathbf{o_n}$ and transform them with an array of fully-connected layers into a vector $\mathbf{o_t}$. We concatenate $\mathbf{o_t}$ with the summary vector $\mathfrak{s}$ and apply another fully-connected layer to get the mention detection score $p_n = \sigma(g_n(\mathbf{o_t} \| \mathfrak{s}))$.

\subsection{Global entity assignment}

The first step in our system is extracting all possible overlapping n-grams from the input texts. We assume that each span in the input text can only refer to a single entity and therefore resolve overlaps by computing a global assignment using the model scores for each n-gram (Step 4 in Figure~\ref{fig:system-diagram}). 

If the mention detection score $p_n$ is above the $0.5$-threshold, the n-gram is predicted to be a correct entity mention and the ranking scores $\mathbf{p_c}$ are used to disambiguate it to a single entity candidate. N-grams that have $p_n$ lower than the threshold are filtered out.

We follow \citet{Guo2013b} in computing the global assignment and hence, arrange all n-grams selected as mentions into non-overlapping combinations and use the individual scores $p_n$ to compute the probability of each combination. The combination with the highest probability is selected as the final set of entity mentions. We have observed in practice a similar effect as descirbed by \citet{Strubell2017}, namely that DCNNs are able to capture dependencies between different entity mentions in the same context and do not tend to produce overlapping mentions.

\subsection{Composite Loss Function}
\label{sec:loss}

Our model jointly computes two scores for each n-gram: the mention detection score $p_n$ and the disambiguation score $p_c$. We optimize the parameters of the whole model jointly and use the loss function that combines penalties for the both scores for all n-grams in the input question:
\begin{multline*}
\mathcal{L} = \sum_{n\in N}\sum_{c\in C_n}\mathcal{M}(t_n, p_n)	+ t_n\mathcal{D}(t_c, p_c),
\end{multline*}
where $t_n$ is the target for mention detection and is either $0$ or $1$, $t_c$ is the target for disambiguation and ranges from $0$ to the number of candidates $|C|$. 

For the mention detection loss $\mathcal{M}$, we include a weighting parameter $\alpha$ for the negative class as the majority of the instances in the data are negative:
\begin{multline*}
\mathcal{M}(t_n,p_n) = - t_n\log p_n - \alpha(1-t_n)\log(1-p_n)
\end{multline*}

The disambiguation detection loss $\mathcal{D}$ is a maximum margin loss:
\begin{equation*}
\mathcal{D}(t_c, p_c) = \frac{\sum_{i=0}^{|C|} \max(0, (m - p_c[t_c] + p_c[i]))}{|C|},
\end{equation*}
where $m$ is the margin value. We set $m=0.5$, whereas the $\alpha$ weight is optimized with the other hyper-parameters.

\subsection{Architecture comparison} 

Our model architecture follows some of the ideas presented in \citet{Francis-Landau2016}: they suggest computing a similarity score between an entity and the context for different context granularities. \citet{Francis-Landau2016} experiment on entity linking for Wikipedia and news articles and consider the word-level and document-level contexts for entity disambiguation. As described above, we also incorporate different context granularities with a number of key differences: 
\begin{enumerate*}[label=(\arabic*)]
  \item we operate on sentence level, word level and character level, thus including a more fine-grained range of contexts;
  \item the knowledge base contexts that \citet{Francis-Landau2016} use are the Wikipedia title and the article texts --- we, on the other hand, employ the structure of the knowledge base and encode relations and related entities;
  \item \citet{Francis-Landau2016} separately compute similarities for each type of context, whereas we mix them in a single end-to-end architecture;
  \item we do not rely on manually defined features in our model.
\end{enumerate*}

\section{Datasets}

\begin{table}[t]
  \begin{center}
  \begin{tabular}{p{0.45\linewidth}
  >{\raggedleft}p{0.2\linewidth}
  >{\raggedleft\arraybackslash}p{0.2\linewidth}}
  \toprule 
  & \#Questions & \#Entities \\ 
   \midrule 
 WebQSP Train & 3098 &  3794\\
 WebQSP Test & 1639 & 2002\\
 \midrule 
 GraphQuestions Test & 2608 & 4680 \\
  \bottomrule
  \end{tabular} 
  \end{center}
  \caption{Dataset statistics \label{table:dataset-stats}}
\end{table}

We compile two new datasets for entity linking on questions that we derive from publicly available question answering data: WebQSP~\cite{Yih2016} and GraphQuestions~\cite{Su2016}.

WebQSP contains questions that were originally collected for the WebQuestions dataset from web search logs~\cite{Berant2013}. They were manually annotated with SPARQL queries that can be executed to retrieve the correct answer to each question. Additionally, the annotators have also selected the main entity in the question that is central to finding the answer. The annotations and the query use identifiers from the Freebase knowledge base.

We extract all entities that are mentioned in the question from the SPARQL query. For the main entity, we also store the correct span in the text, as annotated in the dataset. In order to be able to use Wikidata in our experiments, we translate the Freebase identifiers to Wikidata IDs.

The second dataset, GraphQuestions, was created by collecting manual paraphrases for automatically generated questions~\cite{Su2016}. The dataset is meant to test the ability of the system to understand different wordings of the same question. In particular, the paraphrases include various references to the same entity, which creates a challenge for an entity linking system. 
 The following are three example questions from the dataset that contain a mention of the same entity:
 \ex. \a. \label{ex:graphquestions} what is the rank of marvel's \textbf{iron man}?
 \b. \textbf{iron-man} has held what ranks?
 \b. \textbf{tony stark} has held what ranks?

GraphQuestions does not contain main entity annotations, but includes a SPARQL query structurally encoded in JSON format. The queries were constructed manually by identifying the entities in the question and selecting the relevant KB relations.  We extract gold entities for each question from the SPARQL query and map them to Wikidata. 

We split the WebQSP training set into train and development subsets to optimize the neural model. We use the GraphQuestions only in the evaluation phase to test the generalization power of our model. 
The sizes of the constructed datasets in terms of the number of questions and the number of entities are reported in Table~\ref{table:dataset-stats}. In both datasets, each question contains at least one correct entity mention.

\section{Experiments}

\begin{table}[t]
  \begin{center}
    \begin{tabular}{>{\raggedleft}p{0.4\linewidth}
      >{\raggedleft}p{0.1\linewidth}
      >{\raggedleft}p{0.1\linewidth}   
      >{\raggedleft\arraybackslash}p{0.1\linewidth}}
    \toprule
    & P & R & F1\\ 
    \midrule 
Heuristic baseline  &  \num{0.28591256072172105} &  \num{0.62141779788838614} &  \num{0.39163498098859323} \\
  Simplified VCG & \num{0.80360065466448449} & \num{0.65379494007989347} & \num{0.72099853157121874} \\
  \textbf{VCG}  & \num{0.8234295415959253} & \num{0.64580559254327563} & \num{0.72388059701492524} \\ 
  \bottomrule
  \end{tabular} 
  \end{center}
  \caption{Evaluation results on the \textsc{WebQSP} development dataset (all entities) \label{table:eval-webqsp-dev}}
\end{table}

\subsection{Evaluation Methodology}

We use precision, recall and F1 scores to evaluate and compare the approaches. We follow \citet{Carmel2014} and \citet{Yang2015a} and define the scores on a per-entity basis. Since there are no mention boundaries for the gold entities, an extracted entity is considered correct if it is present in the set of the gold entities for the given question. We compute the metrics in the micro and macro setting. The macro values are computed per entity class and averaged afterwards. 

For the WebQSP dataset, we additionally perform a separate evaluation using only the information on the main entity. The main entity has the information on the boundary offsets of the correct mentions and therefore for this type of evaluation, we enforce that the extracted mention has to overlap with the correct mention. QA systems need at least one entity per question to attempt to find the correct answer. Thus, evaluating using the main entity shows how the entity linking system fulfills this minimum requirement.

\subsection{Baselines}

\textbf{Existing systems}
In our experiments, we compare to DBPedia Spotlight that was used in several QA systems and represents a strong baseline for entity linking\footnote{We use the online end-point: \url{http://www.dbpedia-spotlight.org/api}}. In addition, we are able to compare to the state-of-the-art S-MART system, since their output on the WebQSP datasets was publicly released\footnote{\url{https://github.com/scottyih/STAGG}}. The S-MART system is not openly available, it was first trained on the NEEL 2014 Twitter dataset and later adapted to the QA data~\citep{Yih2015}. 

We also include a heuristics baseline that ranks candidate entities according to their frequency in Wikipedia. This baseline represents a reasonable lower bound for a Wikidata based approach.

\textbf{Simplified VCG}
To test the effect of the end-to-end context encoders of the VCG network, we define a model that instead uses a set of features commonly suggested in the literature for EL on noisy data. In particular, we employ features that cover 
\begin{enumerate*}[label=(\arabic*)]
  \item frequency of the entity in Wikipedia,
  \item edit distance between the label of the entity and the token n-gram,
  \item number of entities and relations immediately connected to the entity in the KB,
  \item word overlap between the input question and the labels of the connected entities and relations,
  \item length of the n-gram.
\end{enumerate*}
We also add an average of the word embeddings of the question tokens and, separately, an average of the embeddings of tokens of entities and relations connected to the entity candidate. We train the simplified VCG model by optimizing the same loss function in Section~\ref{sec:loss} on the same data.

\subsection{Practical considerations}

 \begin{table}[t]
   \begin{center}
   \begin{tabular}{>{\raggedleft}p{0.045\linewidth}
   >{\raggedleft}p{0.045\linewidth}
   >{\raggedleft}p{0.045\linewidth}
   >{\raggedleft}p{0.045\linewidth}
   >{\raggedleft}p{0.045\linewidth}  
   >{\raggedleft}p{0.14\linewidth}
   >{\raggedleft}p{0.14\linewidth}
   >{\raggedleft\arraybackslash}p{0.045\linewidth}}
   \toprule 
   \multicolumn{5}{c}{emb. size} & \multicolumn{2}{c}{filter size} &\\
   $d_w$ & $d_z$  & $d_e$ & $d_r$ & $d_p$ & $\mathbf{DCNN}_w$ & $\mathbf{DCNN}_c$ & $\alpha$\\
   \midrule
   $50$ & $25$  & $50$ & $50$ & $5$ & $64$ & $64$ & $0.5$\\
   \bottomrule
   \end{tabular} 
   \end{center}
   \caption{Best configuration for the VCG model \label{table:hyper-params}}
 \end{table}

\begin{table*}[t]
  \begin{center}
  \begin{tabular}{>{\raggedleft}p{0.18\linewidth}
  >{\raggedleft}p{0.09\linewidth}
  >{\raggedleft}p{0.05\linewidth}
  >{\raggedleft}p{0.05\linewidth}
  >{\raggedleft}p{0.09\linewidth}
  >{\raggedleft}p{0.05\linewidth}
  >{\raggedleft}p{0.05\linewidth}
  >{\raggedleft}p{0.07\linewidth}
  >{\raggedleft}p{0.05\linewidth}
  >{\raggedleft\arraybackslash}p{0.05\linewidth}}
  \toprule
  & \multicolumn{3}{c}{Main entity} & \multicolumn{6}{c}{All entities} \\  
   & P & R & F1 & P & R & F1 & \texttt{m}P & \texttt{m}R & \texttt{m}F1\\ 
   \midrule 
   DBPedia Spotlight  & \num{0.6678082191780822} & \num{0.5948749237339841} & \num{0.6292352371732818} & \num{0.7054794520547946} & \num{0.5144855144855145} & \num{0.5950317735413055} & \num{0.5719900778946438} & \num{0.39161458760733775} & \num{0.451672336820253} \\
   S-MART & \num{0.6343669250645995} & \textbf{\num{0.8987187309334961}} & \num{0.7437515778843726}  & \num{0.6658053402239449} & \textbf{\num{0.7722277722277723}} & \num{0.7150786308973174} & \num{0.6070382176542434} & \textbf{\num{0.6098061685153022}}  & \num{0.5505615532942725} \\
   \midrule
  Heuristic baseline & \num{0.28227408142999005} & \num{0.6937156802928615} & \num{0.40127051349920595} & \num{0.30213505461767626} & \num{0.6078921078921079} & \num{0.4036484245439469} & \num{0.32978864375462974} & \num{0.5372179597771176} & \num{0.3778898456293024} \\
  Simplified VCG & \textbf{\num{0.804040404040404}} & \num{0.7284929835265406} & \num{0.764404609475032} & \textbf{\num{0.837037037037037}} & \num{0.6208791208791209} & \num{0.7129337539432177} & \num{0.658857188648659} & \num{0.4941415444752956} & \num{0.5462362596886855} \\ 
  \textbf{VCG}  & \num{0.7934301958307012} & \num{0.7663209273947529} & \underline{\textbf{\num{0.7796399751707014}}}  & \num{0.8262792166771952} & \num{0.6533466533466533} & \underline{\textbf{\num{0.7297071129707112}}} & \textbf{\num{0.6759776711454646}} & \num{0.5192149330859581} & \underline{\textbf{\num{0.5676951487707222}}} \\ 
  \bottomrule
  \end{tabular} 
  \end{center}
  \caption{Evaluation results on the \textsc{WebQSP} test dataset, the \texttt{m} prefix stands for \textit{macro} \label{table:eval-webqsp-test}}
\end{table*}

\begin{table}[t]
  \begin{center}
  \begin{tabular}{>{\raggedleft}p{0.4\linewidth}
  >{\raggedleft}p{0.18\linewidth}
  >{\raggedleft}p{0.1\linewidth}   
  >{\raggedleft\arraybackslash}p{0.1\linewidth}}
  \toprule 
  & P & R & F1\\ 
    \midrule 
    DBPedia Spotlight & \num{0.38591342626559061} & \num{0.45325290822921155} & \num{0.41688131563304931} \\
    \textbf{VCG} & \textbf{\num{0.58937544867193103}} & \num{0.35372684187850062} & \underline{\textbf{\num{0.44211093161012383}}}\\ 
  \bottomrule
  \end{tabular} 
  \end{center}
  \caption{Evaluation results on \textsc{GraphQuestions}\label{table:eval-graph-test}}
\end{table}

The hyper-parameters of the model, such as the dimensionality of the layers and the size of embeddings, are optimized with random search on the development set. The model was particularly sensitive to tuning of the negative class weight $\alpha$ (see Section~\ref{sec:loss}). Table~\ref{table:hyper-params} lists the main selected hyper-parameters for the VCG model\footnote{The complete list of hyper-parameters and model characteristics can be found in the accompanying code repository.} and we also report the results for each model's best configuration on the development set in Table~\ref{table:eval-webqsp-dev}.

\subsection{Results}

Table~\ref{table:eval-webqsp-test} lists results for the heuristics baseline, for the suggested Variable Context Granularity model (VCG) and for the simplified VCG baseline on the test set of WebQSP.
The simplified VCG model outperforms DBPedia Spotlight and achieves a result very close to the S-MART model. 
Considering only the main entity, the simplified VCG model produces results better than both DBPedia Spotlight and S-MART. 
The VCG model delivers the best F-score across the all setups. We observe that our model achieves the most gains in precision compared to the baselines and the previous state-of-the-art for QA data. 

VCG constantly outperforms the simplified VCG baseline that was trained by optimizing the same loss function but uses manually defined features. Thereby, we confirm the advantage of the mixing context granularities strategy that was suggested in this work.
Most importantly, the VCG model achieves the best macro result which indicates that the model has a consistent performance on different entity classes.

We further evaluate the developed VCG architecture on the GraphQuestions dataset against the DBPedia Spotlight. We use this dataset to evaluate VCG in an out-of-domain setting: neither our system nor DBPedia Spotlight were trained on it. The results for each model are presented in Table~\ref{table:eval-graph-test}. We can see that GraphQuestions provides a much more difficult benchmark for EL. The VCG model shows the overall F-score result that is better than the DBPedia Spotlight baseline by a wide margin. It is notable that again our model achieves higher precision values as compared to other approaches and manages to keep a satisfactory level of recall.

\begin{figure}[tb]
  \begin{tikzpicture}
  \begin{axis}[
      ybar,
    height=7cm,
    width=\linewidth,    
      enlarge y limits=0.15,
      bar width=0.75ex,
      legend style={at={(0.5,-0.25)},
        anchor=north,legend columns=-1},
      ylabel={F-score per class},
      symbolic x coords={event,location,organization,person,fic. character,product,thing,professions etc.},
      xtick=data,
    ymin=0.2,
    nodes near coords,
    nodes near coords align={vertical},
    every node near coord/.append style={font=\tiny, /pgf/number format/fixed,rotate=90,anchor=west},
      x tick label style={rotate=30,anchor=east,font=\small,text width=2cm, align=right},
      y tick label style={font=\small}, 
      ]
      \addplot[cb_darkgreen!80,fill=cb_lightgreen!50] coordinates {
        (fic. character, 0.7272727272727273)
        (event, 0.2576687116564417)
        (location, 0.8166666666666665)
        (organization, 0.6171428571428571)
        (professions etc., 0.26086956521739135)
        (person, 0.9513157894736842)
        (product, 0.5023696682464455)
        (thing, 0.27118644067796605)
        };      
  \addplot[cb_darkblue,fill=cb_blue,pattern=north west lines,pattern color=cb_blue] coordinates {
  (fic. character, 0.6714285714285714)
  (event, 0.3282051282051282)
  (location, 0.74235807860262)
  (organization, 0.7545454545454545)
  (professions etc., 0.1345565749235474)
  (person, 0.9136986301369862)
  (product, 0.543046357615894)
  (thing, 0.282051282051282)
  };
   \addplot[cb_darkyellow,fill=cb_yellow,pattern=crosshatch,pattern color=cb_darkyellow] coordinates {(fic. character, 0.6511627906976745)
   (event,  0.3626373626373627)
   (location, 0.7854984894259819)
   (organization, 0.7416666666666667)
   (professions etc., 0.1377245508982036)
   (person, 0.9106901217861977)
   (product, 0.5771812080536913)
   (thing, 0.3755)
  };  
  \legend{S-MART,Simplified VCG, VCG}
  \end{axis}
  \end{tikzpicture}
  \caption{Performance accross entity classes on \textsc{WebQSP} test dataset \label{fig:eval-webqsp-types}} 
  \end{figure}
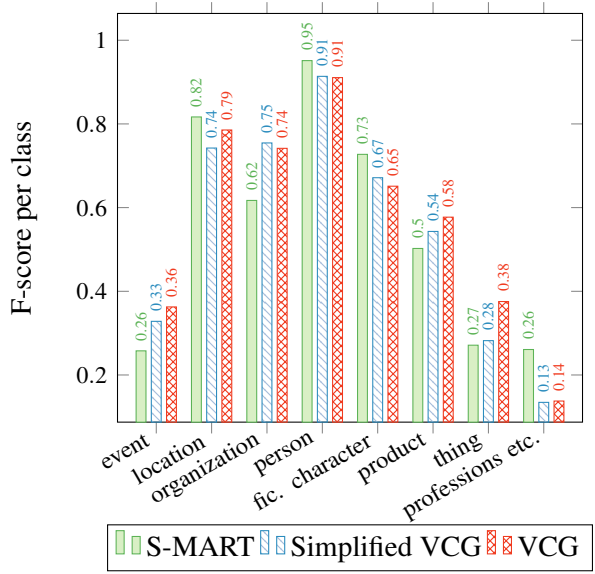

\begin{table*}[t]
  \begin{center}
    \begin{tabular}{>{\raggedleft}p{0.24\linewidth}
      >{\raggedleft}p{0.07\linewidth}
      >{\raggedleft}p{0.05\linewidth}
      >{\raggedleft}p{0.05\linewidth}
      >{\raggedleft}p{0.07\linewidth}
      >{\raggedleft}p{0.05\linewidth}
      >{\raggedleft}p{0.05\linewidth}
      >{\raggedleft}p{0.06\linewidth}
      >{\raggedleft}p{0.05\linewidth}
      >{\raggedleft\arraybackslash}p{0.05\linewidth}}
      \toprule
      & \multicolumn{3}{c}{Main entity} & \multicolumn{6}{c}{All entities} \\  
       & P & R & F1 & P & R & F1 & \texttt{m}P & \texttt{m}R & \texttt{m}F1\\ 
       \midrule 
       \textbf{VCG}  & \num{0.7934301958307012} & \textbf{\num{0.7663209273947529}} & \underline{\textbf{\num{0.7796399751707014}}}  
       & \textbf{\num{0.8262792166771952}} & \textbf{\num{0.6533466533466533}} & \underline{\textbf{\num{0.7297071129707112}}} 
       & \textbf{\num{0.6759776711454646}} & \textbf{\num{0.5192149330859581}} & \underline{\textbf{\num{0.5676951487707222}}} \\
        w/o token context & \num{0.7822950819672131} & \num{0.7278828553996339} & \num{0.7541087231352719}  
        & \num{0.8118032786885246} & \num{0.6183816183816184} & \num{0.7020130422455344} 
       	& \num{0.66386029} & \num{0.47420068} & \num{0.52970676} \\
        w/o character context & \textbf{\num{0.801859799713877}} & \num{0.6839536302623551} & \num{0.73822851498189}  
        & \num{0.8204577968526466} & \num{0.5729270729270729} & \num{0.6747058823529412} 
       	& \num{0.66660459} & \num{0.40368901} & \num{0.47147374} \\
        w/o KB structure context & \num{0.701765447667087} & \num{0.6790726052471019} & \num{0.6902325581395349}  
        & \num{0.7276166456494325} & \num{0.5764235764235764} & \num{0.6432552954292086} 
       	& \num{0.54878382} & \num{0.42728859} & \num{0.46127831} \\
        w/o KB lexical context & \num{0.7826370757180157} & \num{0.7315436241610739} & \num{0.7562283191422265}  
        & \num{0.8067885117493473} & \num{0.6173826173826173} & \num{0.6994906621392191} 
       	& \num{0.6427728} & \num{0.45399265} & \num{0.50846635} \\ 
      \bottomrule
      \end{tabular} 
  \end{center}
  \caption{Ablation experiments for the VCG model on \textsc{WebQSP}\label{table:ablation-webqsp-test}}
\end{table*}

\textbf{Analysis} In order to better understand the performance difference between the approaches and the gains of the VCG model, we analyze the results per entity class (see Figure~\ref{fig:eval-webqsp-types}). We see that the \mbox{S-MART} system is slightly better in the disambiguation of Locations, Person names and a similar category of Fictional Character names, while it has a considerable advantage in processing of Professions and Common Nouns. Our approach has an edge in such entity classes as Organization, Things and Products. The latter category includes movies, book titles and songs, which are particularly hard to identify and disambiguate since any sequence of words can be a title. VCG is also considerably better in recognizing Events. We conclude that the future development of the VCG architecture should focus on the improved identification and disambiguation of professions and common nouns.

To analyze the effect that mixing various context granularities has on the model performance, we include ablation experiment results for the VCG model (see Table~\ref{table:ablation-webqsp-test}). We report the same scores as in the main evaluation but without individual model components that were described in Section~\ref{sec:architecture}.

We can see that the removal of the KB structure information encoded in entity and relation embeddings results in the biggest performance drop of almost 10 percentage points. The character-level information also proves to be highly important for the final state-of-the-art performance. These aspects of the model (the comprehensive representation of the KB structure and the character-level information) are two of the main differences of our approach to the previous work. Finally, we see that excluding the token-level input and the lexical information about the related KB relations also decrease the results, albeit less dramatically.

\section{Conclusions}

We have described the task of entity linking on QA data and its challenges. The suggested new approach for this task is a unifying network that models contexts of variable granularity to extract features for mention detection and entity disambiguation. This system achieves state-of-the-art results on two datasets and outperforms the previous best system used for EL on QA data. The results further verify that modeling different types of context helps to achieve a better performance across various entity classes (macro f-score). 

Most recently, \citet{Peng2017} and \citet{Yu2017} have attempted to incorporate entity linking into a QA model. This offers an exciting future direction for the Variable Context Granularity model.

\section*{Acknowledgments}
This work has been supported by the German Research Foundation as part of the Research Training Group Adaptive Preparation of Information from Heterogeneous Sources (AIPHES) under grant No. GRK 1994/1, and via the QA-EduInf project (grant GU 798/18-1 and grant RI 803/12-1). 

We gratefully acknowledge the support of NVIDIA Corporation with the donation of the Titan X Pascal GPU used for this research.

\bibliography{entity-linking}
\bibliographystyle{acl_natbib}

\end{document}